\documentclass[10pt, a4paper]{article}

\usepackage{lrec-coling2024} 

\usepackage{listings}
\usepackage{multirow}
\usepackage{multicol}
\usepackage{natbib}
\usepackage{multibib}
\usepackage{hyperref}
\usepackage{cleveref}
\usepackage{fontawesome5}
\makeatletter
\def\@mb@citenamelist{cite,citep,citet,citealp,citealt,citepalias,citetalias}
\makeatother
\newcites{languageresource}{~}

\usepackage{graphicx}
\usepackage{subcaption}
\usepackage{tabularx}
\usepackage{soul}
\usepackage{amssymb}
\usepackage{titlesec}
\usepackage{xcolor}
\definecolor{darkblue}{rgb}{0, 0, 0.5}
\hypersetup{colorlinks=true, citecolor=darkblue, linkcolor=darkblue, urlcolor=darkblue}
\usepackage{xstring}
\usepackage{color}
\usepackage{pifont}
\usepackage[normalem]{ulem}

\newcommand{\ourcheck}{\ding{51}}
\newcommand{\ourcross}{\ding{53}}
\newcommand{\corpus}[0]{\textsc{BarNER}}

\usepackage{scalerel}
\NewDocumentCommand\emojigermany{}{\scalerel*{\includegraphics{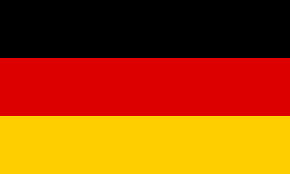}}{X}}

\title{\textit{Sebastian, Basti, Wastl?!} \\ Recognizing Named Entities in Bavarian Dialectal Data}

\name{Siyao Peng\textsuperscript{\faMountain\kern1pt\faRobot}  \quad
Zihang Sun\textsuperscript{\faMountain}\quad
Huangyan Shan\textsuperscript{\faMountain}\quad
Marie Kolm\textsuperscript{\faMountain}
\\
\large\bf 
Verena Blaschke\textsuperscript{\faMountain\kern1pt\faRobot}  \quad
Ekaterina Artemova\textsuperscript{\faMountain}\quad
Barbara Plank\textsuperscript{\faMountain\kern1pt\faRobot} 
} 

\address{\textsuperscript{\faMountain} MaiNLP, Center for Information and Language Processing, LMU Munich, Germany \\
\textsuperscript{\faRobot} Munich Center for Machine Learning (MCML), Munich, Germany \\
{\tt \{siyao.peng, b.plank\}@lmu.de}
}

\abstract{
Named Entity Recognition (NER) is a fundamental task to extract key information from texts, but annotated resources are scarce for dialects.
This paper introduces the first dialectal NER dataset for German, \corpus{}, with 161K tokens annotated on Bavarian Wikipedia articles (\textit{bar-wiki}) and tweets (\textit{bar-tweet)}, using a schema adapted from German CoNLL 2006 and GermEval.
The Bavarian dialect differs from standard German in lexical distribution, syntactic construction, and entity information.
We conduct in-domain, cross-domain, sequential, and joint experiments on two Bavarian and three German corpora and present the first comprehensive NER results on Bavarian.
Incorporating knowledge from the larger German NER (sub-)datasets notably improves on  \textit{bar-wiki} and moderately on \textit{bar-tweet}. 
Inversely, training first on Bavarian contributes slightly to the seminal German CoNLL 2006 corpus.
Moreover, with gold dialect labels on Bavarian tweets, we assess multi-task learning between five NER and two Bavarian-German dialect identification tasks and achieve NER SOTA on \textit{bar-wiki}.
We substantiate the necessity of our low-resource \corpus{} corpus and the importance of diversity in dialects, genres, and 
 topics in enhancing model performance.  
 \\ \newline  
}

\begin{document}

\maketitleabstract

\section{Introduction}
Named Entity Recognition (NER) is a long-standing Natural Language Processing (NLP) task that extracts named entities (NEs) from texts and classifies them into a closed set of semantic types.
A large number of NER datasets annotated for different genres and languages emerged after the seminal benchmark CoNLL 2003 shared task 
\citep{tjong-kim-sang-de-meulder-2003-introduction}, such as 
English \citep{baldwin-etal-2015-shared,  strauss-etal-2016-results, derczynski-etal-2017-results, liu-ritter-2023-conll}, 
German \citep{benikova-etal-2014-nosta},
code-switched and multilingual corpora \citep{aguilar-etal-2018-named, piskorski-etal-2017-first, singh-2008-named, liu-2021-crossNER, fetahu-etal-2023-semeval}, to name but a few.
However, NER datasets for non-standard language varieties are scarce --- there remains a demand for high-quality manual annotations on low-resource dialects.

Bavarian (German:\ \textit{Bairisch}; Bavarian:\ \textit{Boarisch}; ISO: \textit{639-3}; code: \textit{bar}) is a West German dialect spoken in southern Germany, Austria, and northern Italy (South Tyrol).
Bavarian has distinctive features in phonology, lexicons, and syntax compared to Standard German \citep{hinderling1984bairisch, rowley2011bavarian}.
Given its large number of speakers (10M+,  \citealt{rowley2011bavarian}), 
and regional variations in writings, we chose Bavarian to exemplify high-quality manual NER annotations on a non-standard language variety. 

\textbf{This paper presents \corpus{}, the first manually annotated NER dataset on a German dialect.}
\corpus{} contains 161K Bavarian tokens in two genres: Wikipedia articles and tweets. 
We include coarse-grained person (PER), location (LOC), organization (ORG), and miscellaneous (MISC) entities strictly mirroring the CoNLL  2006 German guideline \citep{tjong-kim-sang-de-meulder-2003-introduction}, as well as fine-grained annotations on derived and partially-contained NEs and other entity types adapted from GermEval 2014 \citep{reimers2014germeval}. 
We conduct manual double annotations on half of the dataset and achieve high inter-annotator agreements.
Primary and double BarNER annotations are publicly available on Github\footnote{\url{https://github.com/mainlp/BarNER}} as much as licenses allow. 
We also provide a training-centric NE annotation guideline on GitHub.  
We follow \citet{bender-friedman-2018-data} to include a data statement in Appendix \ref{sec:appx-data-statement}.

\textbf{We highlight token- and entity-level distinctions between Bavarian and German NER.}
We compare our Bavarian wiki (\textit{bar-wiki}) and tweet (\textit{bar-tweet}) data to two German datasets within the same genres and the  CoNLL 2006 German news dataset to illustrate how lexically distinct Bavarian is from German.
We find cross-dialectal lexical dissimilarities to be larger than cross-genre in German. 
Moreover, the distribution of entity types and texts varies highly across dialects, genres, and topics -- jointly referred to as ``domains'' in this paper.

\textbf{We conduct in-domain, cross-domain, sequential, and joint NER experiments between Bavarian and German and across genres.}
We establish baseline NER scores on \corpus{} and demonstrate that directly applying German-trained models achieves poor performance on Bavarian.
Experiments show noticeable improvements in Bavarian by sequential and joint training and incorporating knowledge from larger German datasets.
Inversely, training first on Bavarian can also help German NER, though to a smaller extent. 

\textbf{For multi-task learning (MTL), we train NER with Bavarian-German dialect identification.}
MTL scores SOTA on \textit{bar-wiki} NER, 11.26 points higher on Span F1 than the in-domain baseline. 
Results demonstrate the efficacy of our multi-genre dialectal data in establishing low-resource evaluations and advancing high-resource performances.

\section{Related Work}\label{sec:related-work}

\paragraph{NER datasets for German}\label{subsec:related-work-ner-german}

The CoNLL 2003 shared task  \citep{tjong-kim-sang-de-meulder-2003-introduction} provides seminal NER datasets for German and English.
Annotation guidelines and the German CoNLL dataset are subsequently updated to CoNLL 2006.
On the other hand, German  NoSta-D (Non-Standard German, \citealt{dipper2013nosta}) NER annotations were first implemented on five non-standard genre varieties -- historical data, chat data, spoken data, learner data,
and literary prose -- and later extended to the prominent GermEval 2014 shared task \citep{benikova-etal-2014-nosta} on Wikipedia and online news.
GermEval 2024 annotates nested NEs (though limited to two layers), and the \makebox{-part} and \makebox{-deriv} suffixes for compounds that partly contain NEs and words morphologically derived from NEs. 
Moreover, German NER datasets were expanded to various domains, such as 
parliament debates \citep{faruqui2010training},
historical newspapers \citep{neudecker-2016-open, hamdi-2021-newseye},
historical biodiversity literature (BIOfid, \citealt{ahmed-etal-2019-biofid}),
biographic interviews \citep{ruppenhofer-etal-2020-fine},
legal court decisions \citep{leitner-etal-2020-dataset}, 
and traffic reports \citep{schiersch-etal-2018-german, hennig-etal-2021-mobie}.
More fine-grained entity types are introduced in these corpora.


\paragraph{NER datasets for dialects and low-resource languages}\label{subsec:related-work-ner-dialects}

The few dialectal NER datasets are predominately annotated for Arabic, where most surfaced in the last two years. 
These include Egyptian Arabic news and blogs \citep{darwish-2013-named, zirikly-diab-2014-named},
Palestinian Arabic social media texts (Wojood, \citealt{jarrar-etal-2022-wojood}),
Narabizi (Latin-scripted vernacular Arabic) forums  (NERDz, \citealt{touileb-2022-nerdz}),
Algerian Facebook pages and YouTube channels (DzNER, \citealt{dahou-2023-DzNER}),
and Darija (Moroccan vernacular Arabic) Wikipedia pages (DarNERcorp, \citealt{moussa-2023-DarNERcorp}).
\citet{orasmaa-etal-2022-named} annotate Parish court records for north and south Estonian dialects (Finnic family).
Besides, NER datasets emerged for low-resource languages, such as
Assamese \citep{pathak-etal-2022-asner},
Marathi \citep{litake-etal-2022-l3cube},
Kazakh \citep{yeshpanov-etal-2022-kaznerd},
Vietnamese \citep{phan-etal-2022-named},
and Sub-Saharan languages \citep{adelani-etal-2022-masakhaner}.

Beyond gold annotations, \citet{pan-etal-2017-cross} create a silver WikiAnn dataset for 282 languages through Wikipedia knowledge mining and cross-lingual transfer, which includes Bavarian.
However, many sentences in the small or medium-sized sub-corpora are unnatural. 
For example, the average sentence length is low ($\sim$5.26 tokens per sentence) in Bavarian, and half are hyperlinks,
e.g., \textit{Deitschland `[Deitschland]\textsubscript{LOC}'} (the word for `Germany' in Bavarian).
Thus, there remains a gap for low-resource but high-quality NER datasets.

\paragraph{NER datasets for tweets}\label{subsec:related-work-ner-tweet}
Tweets form an unparalleled informal genre in NLP research given its (previously) short length limit, many users, and most timely posts \citep{finin-etal-2010-annotating, scanner-2022-social-media-tweet}. 
Tweets contain NEs in texts, \#hashtags, @mentions, and emojis.
\citet{finin-etal-2010-annotating} crowd-sourced the earliest Twitter NER dataset 
for English. 
\citet{ritter-etal-2011-named} ignored @mentions due to ambiguity and anonymity, and the corpus was used 
in Workshop on Noisy User-generated Text (WNUT) 2015-2017 shared tasks \citep{baldwin-etal-2015-shared, strauss-etal-2016-results, derczynski-etal-2017-results}. 
The Broad Twitter Corpus \citep{derczynski-etal-2016-broad}
samples from heterogeneous temporal, geographical, and social contexts.
Tweebank-NER \citep{jiang-etal-2022-annotating} adds NEs to Universal Dependencies Tweebank V2 \citep{liu-etal-2018-parsing},
and TweetNER7 \citep{ushio-etal-2022-named} is the largest tweet NER corpus to date.
Beyond English, tweet NEs are annotated for 
German \citep{schiersch-etal-2018-german, hennig-etal-2021-mobie}, 
French \citep{lopez2017cap}, 
Danish \citep{plank-etal-2020-dan},  
Turkish \citep{küçük2019tweet, carik-yeniterzi-2022-twitter},
Serbian and Croatian \citep{baksa2014named, Baksa_Golović_Glavaš_Šnajder_2017, ljubesc-2019-croatian-tweet}, etc. 

\section{\corpus{}}\label{sec:barner}
\subsection{NER Taxonomy}\label{subsec:ner-taxonomy}
We conduct NER annotations on Bavarian (\textit{bar}) mirroring the CoNLL 2006 \citep{tjong-kim-sang-de-meulder-2003-introduction} and GermEval 2014 (NoSta-D, \citealt{benikova-etal-2014-nosta}) guidelines for German. 
We constrain our annotations to the narrowly defined but widely adapted flat NEs compatible with sequential BIO tagging. Namely, we exclude common nouns, pronouns, overlapping, or nested NEs. 

We follow both guidelines to include the four major entity types: PER, ORG, LOC, and MISC. 
Since nominal derivation and compounding are similarly prevalent in Bavarian, we adapt NoSta-D's strategy to add \makebox{-deriv} and \makebox{-part} suffixes for tokens derived or partly containing NEs. 
For example, [\textit{Italienroas}]\textsubscript{LOCpart} `tour of Italy' partly contains the country Italy and [\textit{eiropäischn}]\textsubscript{LOCderiv} `European' is an adjective derived from [\textit{Eiropa}]\textsubscript{LOC} `Europe.'
Moreover, annotators observe a relatively high frequency of NEs referring to languages (LANG), religions (RELIGION), events (EVENT), and works of art (WOA) during training, and we thus add them and their \makebox{-deriv} and \makebox{-part} suffixed variations to our NE tagset.
These additional labels elaborate on entities in a given text and can be merged or discarded when comparing with other datasets (see \S\ref{subsec:comparison-type-mapping-distributions}).
We refer to the PER/LOC/ORG/MISC tagset as coarse-grained and the extension with \makebox{-part}, \makebox{-deriv}, and other entity types as fined-grained.

\subsection{Genre \& Corpus Statistics}\label{subsec:genre-corpus-stats}
We conduct manual annotations on two mainstream genres:
Wikipedia articles and Twitter (X) tweets. 
Wikipedia pages are carefully written and consistently updated by multiple contributors. 
We selected 40+ documents from Bavarian Wiki\footnote{\url{https://bar.wikipedia.org/wiki/Wikipedia:Hoamseitn}} with a wide topic coverage.
Continuous sections of $\sim$1.5K tokens from the beginning of documents were extracted to enable future document-level analyses.

Social media texts like tweets are noisier, less formal, and more dynamic \citep{ushio-etal-2022-named}.
Tweet collection for Bavarian is also more difficult.
To sample enough data with author diversity, we snowballed from a list of 17 Bavarian `seed users'\footnote{\url{http://indigenoustweets.com/bar/}} \citep{bhroin_social_2015} to their friend circles on Twitter (i.e., the people they follow and follow them), under the assumption that dialect groups are closely connected on social media \citep{backstrom-2006-social-group}.
All tweets\footnote{4.4K+ tweets from 151 users were collected between Feb and May 2023, with no restriction on when the tweets were posted.} from seed users and their friends are extracted to train a pilot dialect identifier to filter these users.
We update the list of seed users iteratively until we reach 100K+ tokens on silver Bavarian tweets. 

We further ensure the dialectal sanity of our tweet data by asking annotators to classify tweet sentences into one of the following categories:
\textit{bar} if the sentence is predominately Bavarian,
\textit{de} if predominately German, 
\textit{other} if another language or dialect, 
and \textit{na} if unintelligible\footnote{Cohen's kappa on intermediate and final dialect identifications are 82.47 and 85.66 between two annotators.} 
-- see \S\ref{sec:multi-task-dialect-identification} for our multi-task learning experiments combining NER with Dialect Identification (DID).
As for our \textit{bar-tweet} NER dataset, we only kept \textit{bar}-labelled sentences.
During NER annotations, we include hashtags ([\textit{\#minga}]\textsubscript{LOC} `Munich') and emojis ([\emojigermany]\textsubscript{LOC}) as our annotation targets.
Mentions are anonymized to \textit{@mention} and excluded from annotation.

Table~\ref{tab:corpus_statistics} presents the number of tokens, sentences, and named entities in our two Bavarian NER sub-corpora: \textit{bar-wiki} and \textit{bar-tweet}.
Both genres reach 75K+ tokens, a quarter the size of the German CoNLL 2006 corpus. 
However, we note that \textit{bar-tweet} has proportionally much fewer entities than \textit{bar-wiki} due to informality and tweet length limits (see  \S\ref{sec:compare-german} for comparisons with German).

\begin{table}[htb]
\centering
\resizebox{0.45\textwidth}{!}{
\begin{tabular}{c|rrrr}
Corpus & \#Toks & \#Sents & \#Ents & \%Ents/Toks \\
\hline
\textit{bar-wiki} & 75,687 & 3,574 & 4,192 & 5.54\% \\
\textit{bar-tweet} & 86,090 & 7,459 & 2,486 & 2.89\% \\
\end{tabular}
}
\caption{\textit{bar-wiki} and \textit{bar-tweet} NER corpus statistics: numbers of tokens, sentences, entities, and percentages of entities over tokens. }
\label{tab:corpus_statistics}
\end{table}

\subsection{Annotation Procedure \& Agreement}\label{subsec:annotation-procedure-agreement}
The annotation project took five months 
and was conducted by three graduate students with computational linguistics backgrounds.\footnote{The annotators were hired and compensated for their work following national salary rates.} 
These include one native Bavarian speaker and two majoring in German studies.
For training, three annotators annotate the same documents independently during the first three months.
Project coordinators hold discussion sessions biweekly and adjudicate the final version with annotators. 
After training, two annotators remained on the project and worked on different documents.
We conduct two inter-annotator agreement (IAA) experiments (i.e., after training and final) between the two remaining annotators on the coarse- and fine-grained labels.
Each IAA is evaluated on $\sim$7K tokens, aligning with NER experiments' development and test sets in \S\ref{sec:ner-experiments-results}.
Table~\ref{tab:ner-inter-annotator-agreement} presents the token- and entity-level IAA statistics. 

\begin{table}[htb]
\centering
\resizebox{0.5\textwidth}{!}{
\begin{tabular}{lcc|rr|rr}
\multirow{2}{*}{Corpus} 
& \multirow{2}{*}{\#Toks}
& \multirow{2}{*}{Tagset}
& \multicolumn{2}{c|}{Token-level} 
& \multicolumn{2}{c}{Entity-level F1} \\
& & & Raw & Kappa & Untyped &  Typed \\
\hline
\multicolumn{7}{c}{Intermediate Agreement (Dev Set)} \\
\multirow{2}{*}{bar-wiki} &  \multirow{2}{*}{7.4K}  
& fine & 97.62 & 88.80 & 91.64 & 83.62  \\
 & &  coarse & 97.93 & 89.76 & 91.82 & 86.13  \\
\multirow{2}{*}{bar-tweet} &  \multirow{2}{*}{7.0K} 
& fine & 99.09 & 87.02 & 86.52 & 83.48  \\
 & &  coarse & 99.29 & 88.19 & 87.21 & 85.64  \\
\hline
\multicolumn{7}{c}{Final Agreement (Test Set)} \\
\multirow{2}{*}{bar-wiki} &  \multirow{2}{*}{6.9K}  
& fine  & 99.11 & 94.04 & 94.99 & 88.32  \\
& &    coarse & 99.24 & 94.64 & 94.84 & 90.52  \\
 \multirow{2}{*}{bar-tweet} &  \multirow{2}{*}{7.3K}  
 & fine  & 99.16 & 88.26 & 89.60 & 85.20  \\
 &  &   coarse & 99.27 & 88.22 & 88.10 & 87.62  \\
\end{tabular}
}
\caption{Inter-Annotator Agreement (IAA) on Bavarian wiki and tweet NER tagging, including token-level raw and Cohen's kappa scores and entity-level untyped and typed F1 scores.}
\label{tab:ner-inter-annotator-agreement}
\end{table}

\noindent
Token- and entity-level agreements on Bavarian wiki and tweet are high; all Cohen's kappa and typed span F1 are above 80, with an additional improvement on the final agreements. 
Since tweet entities are much more sparse than wiki (see Table~\ref{tab:corpus_statistics}), determining whether a noun phrase is an entity span is more difficult for tweets, resulting in a $\sim$5 percentage point decrease in untyped F1.
In contrast, entity typing is easier for tweets with smaller gaps between untyped and typed F1s. 
The IAA scores on coarse- and fine-grained entity types are also close.
Our data release includes individual annotations on top of the adjudicated final version for future annotation disagreement studies.

\section{Comparisons with German}\label{sec:compare-german}

To assess quantitatively and qualitatively how Bavarian NER differs from German, we compare our \corpus{} sub-corpora (\textit{bar-wiki} and \textit{bar-tweet}) with three German ones:
the wiki portion of \makebox{NoSta-D} (henceforth \textit{de-wiki}, \citealt{benikova-etal-2014-nosta}),
~MobIE transportation tweets (\textit{de-tweet}, \citealt{hennig-etal-2021-mobie}),
and the German CoNLL 2006 news (\textit{de-news}, \citealt{tjong-kim-sang-de-meulder-2003-introduction}).
The first two share the same genres as \corpus{}; the last is the benchmark German NER dataset.
\S\ref{subsec:comparison-lexical-similarity} quantifies lexical similarities and \S\ref{subsec:comparison-type-mapping-distributions} maps fine-grained tagsets into coarse-grained for entity type comparisons.
\S\ref{subsec:comparison-entity-texts} presents cross-genre and cross-dialectal distinctions in entity texts, and \S\ref{subsec:comparison-qualitative-observations} supplements with annotators' observations on differences in annotating Bavarian NER compared to German.

\subsection{Lexical Similarities}\label{subsec:comparison-lexical-similarity}

\begin{figure}[h!bt]
\centering
\includegraphics[width=0.45\textwidth]{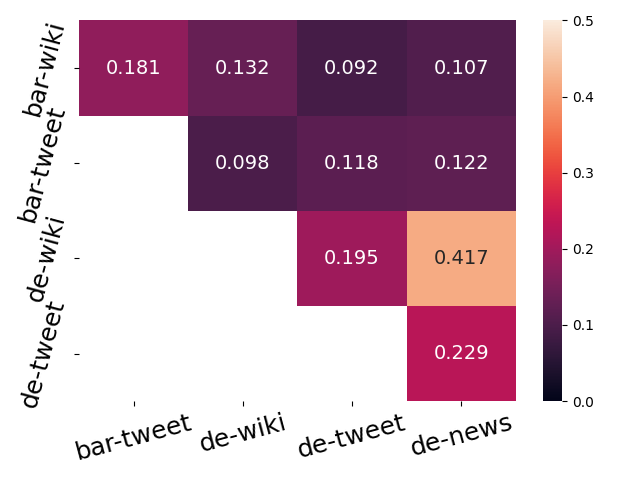}
\caption{Heatmap of Jaccard similarities of top 1K frequent tokens across five (sub-)corpora: \textit{bar-wiki}, \textit{bar-tweet}, \textit{de-wiki}, \textit{de-tweet}, and \textit{de-news}.}
\label{fig:jaccard-heatmap}
\end{figure}

This section demonstrates lexical distinctions from two aspects: 
between orthographically distinctive Bavarian and German;
and among wiki, tweet, and news genres.
We employ Jaccard similarity, which is frequently used in corpus analysis and measures the ratio of shared (i.e., intersection) tokens over concatenated (i.e., union) ones between datasets \citep{chen-etal-2022-semeval}.
Figure \ref{fig:jaccard-heatmap} presents the Jaccard similarities among the top 1K frequent tokens sampled from the Bavarian and German (sub-)corpora 
-- see Appendix \ref{sec:appx-word-cloud} for word clouds.
We value variations in word choice and orthography across dialects and genres and thus compare surface token strings without normalization -- analogous to how language models process texts.

\noindent
Firstly, we observe that German \textit{wiki$\times$news} has the highest lexical similarity of 0.417 due to their formality and well-edited texts.
Secondly, lexical distances between tweets and other same-dialect genres are similar for Bavarian and German, e.g., German \textit{tweet$\times$news} scores 0.229, \textit{tweet$\times$wiki} 0.195, and Bavarian \textit{tweet$\times$wiki} is 0.181.
Lastly, Jaccard similarities between dialects within the same genre are not distinguishable from cross-dialect and cross-genre scenarios. 
For example, \textit{de-wiki$\times$bar-wiki} 0.132 is close to \textit{de-news$\times$bar-tweet} 0.122.
We can report from these observations that Bavarian is more lexically distinct from German than different text genres in German are from each other.

\subsection{Entity Mappings \& Distributions}\label{subsec:comparison-type-mapping-distributions}
The benchmark CoNLL 2003 \citep{tjong-kim-sang-de-meulder-2003-introduction} dataset was updated to CoNLL 2006 with some guideline changes.\footnote{\url{https://usermanual.wiki/Document/guide.820232904.pdf}} 
These include: 
1) adjectives derived from proper names (e.g., \textit{deutsch} `adj.\ German') are no longer marked as NEs; and
2)  neither nouns derived from proper names (e.g., \textit{Frankfurter} `people from Frankfurt')
3) nor nominal compounds containing proper names (e.g., \textit{SPD-Vorsitzender} `SPD-Chairman') are marked.
More recent NER datasets, such as \corpus{}, evolved from CoNLL 2006 and include additional entity types to capture language or domain-specific phenomena.

To conduct objective comparisons, we normalize all fine-grained tagsets to the CoNLL 2006 (\textit{de-news}) coarse-grained one.
Table~\ref{tab:tagset-normalization} shows the tagset normalization rules for the four \textit{bar/de-wiki/tweet} (sub-)datasets. 
These include removing nested, partly, and derived NEs and dropping or merging fine-grained entity types into MISC.

\begin{table}[h!bt]
\centering
\resizebox{0.5\textwidth}{!}{
\begin{tabular}{c|cccc}
Normalization rules
& \begin{tabular}[c]{@{}l@{}}\textit{bar-}\\\textit{wiki}\end{tabular}
& \begin{tabular}[c]{@{}l@{}}\textit{bar-}\\\textit{tweet}\end{tabular}
& \begin{tabular}[c]{@{}l@{}}\textit{de-}\\\textit{wiki}\end{tabular}
& \begin{tabular}[c]{@{}l@{}}\textit{de-}\\\textit{tweet}\end{tabular} \\
\hline
Removing 2nd-level NEs & & & \checkmark & \\
OTH -> MISC & & & \checkmark & \\
EVENT/WOA -> MISC & \checkmark & \checkmark & & \checkmark \\
-part/deriv -> O & \checkmark & \checkmark & \checkmark & \\
LANG/RELIGION -> O &  \checkmark &  \checkmark & & \\
TIME/DISTANCE/NUMBER -> O & & & & \checkmark \\
\end{tabular}
}
\caption{Tagset normalization on Bavarian and German NER (sub-)datasets.}
\label{tab:tagset-normalization}
\end{table}

\begin{figure}[h!bt]
\centering
\includegraphics[width=0.49\textwidth]{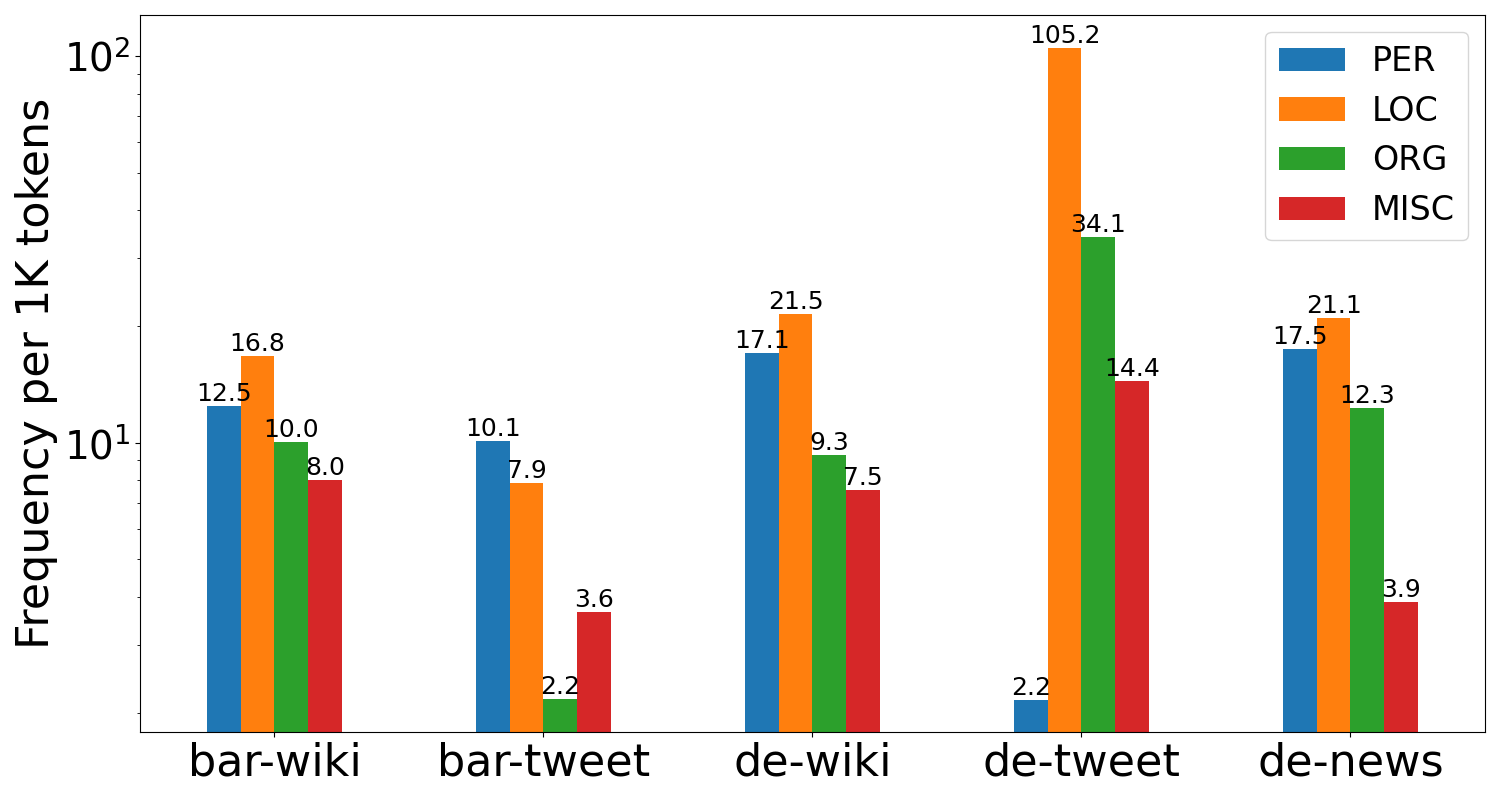}
\caption{Frequencies of named entity types in German and Bavarian (sub-)datasets per 1K tokens (bars are displayed in log-scale).}
\label{fig:freq-type-per-1K-tokens}
\end{figure}

\noindent
Figure \ref{fig:freq-type-per-1K-tokens} presents frequencies of the four coarse-grained entity types per 1K tokens in the five normalized (sub-)datasets.
Firstly, we observe similar proportional frequencies of four entity types in \textit{bar-wiki} and \textit{de-wiki}, whereas \textit{de-news} has a slightly lower ratio for MISC. 
Secondly, \textit{de-tweet} exhibits an extreme outlier of 105.2 on LOC due to the deliberately sampled transportation tweets and the abundance of location entities, such as routes, streets, and cities. 
Lastly, \textit{bar-tweet} has the least amount of entities proportionally, particularly low in ORG.
The higher rank of PER entities than LOC and the frequent use of first-person pronouns in \textit{bar-tweet} suggest that texts in the corpus are more personal than the other four (sub-)datasets, involving fewer proper nouns.
Another contributing factor is the decision to anonymize and ignore all @mentions in annotations; nevertheless, PER is the most frequent NE type in \textit{bar-tweet}. 
Even with normalized NER types, (sub-)datasets vary largely in the distribution of entity annotations, which can contribute positively or negatively when simultaneously or sequentially training on multiple NER (sub-)datasets as in \S\ref{sec:ner-experiments-results}. 

\subsection{Top Entity Texts}\label{subsec:comparison-entity-texts}
Besides the distribution of entity types, we also examine how entity texts differ across genres and dialects. 
Table~\ref{tab:frequent-entities} presents the top 10 entities from the five normalized (sub-)corpora. 
Some entities are dominant and shared across dialects, \textit{Deutschland} `Germany' in German or \textit{Deitschland} in Bavarian. 
The most common cities differ between the two dialects: \textit{Minga} `Munich' is the most popular in Bavarian. In contrast, \textit{Berlin} and \textit{Frankfurt} frequently occur in the German corpora. 
Inevitably, different corpus designs result in the largest divergence in frequencies. 
We constructed our \textit{bar-wiki} sub-corpus to be around 1.5K tokens per document for future discourse analyses, which caused the document titles to be prominent entities.
\textit{bar-tweet} is sampled by a list of Bavarian-speaking users.
Names of their frequently connected friends, e.g., \textit{Michi}, \textit{Gitte}, and \textit{Gela}, are among the top entities in the dataset, particularly occurring in the greeting sentences.
\textit{de-wiki} contains the most city and country names.
In the \textit{de-news} corpus, mentions of the then-currency \textit{Mark} and political parties like \textit{SPD} or \textit{CDU} are prominent.
\textit{de-tweet}, purposely sampled from the transportation domain, is dominated by local railway lines with or without hashtags, e.g., \textit{\#S3} and \textit{S3}.
These different entity distributions across (sub-)datasets contribute to the divergent performances when sequentially or jointly trained on different external NER (sub-)datasets in \S\ref{sec:ner-experiments-results}.

\begin{table}[h!bt]
\centering
\resizebox{0.5\textwidth}{!}{
\begin{tabular}{c|l}
Corpus & Top 10 entities \\
\hline
\textit{bar-wiki} 
& 
\begin{tabular}[c]{@{}l@{}}
Minga\textsubscript{LOC}, 
Odysseus\textsubscript{PER}, 
Nestroy\textsubscript{PER},  
Deitschland\textsubscript{LOC},\\
Bundeswehr\textsubscript{ORG}, 
Dochau\textsubscript{LOC}, 
Los Angeles\textsubscript{LOC}, \\
Google\textsubscript{ORG}, 
Doana\textsubscript{LOC}, 
Bassau\textsubscript{LOC}
\end{tabular} \\
\hline
\textit{bar-tweet} 
& \begin{tabular}[c]{@{}l@{}}
Gela\textsubscript{PER}, 
Michi\textsubscript{PER}, 
Minga\textsubscript{LOC}, 
Gitte\textsubscript{PER}, 
Stefan\textsubscript{PER}, \\   
Jo Jo\textsubscript{PER},  
\#wiesn\textsubscript{MISC}, 
Bayern\textsubscript{LOC}, 
Twitter\textsubscript{ORG}, 
\#Wiesn\textsubscript{MISC}
\end{tabular} \\
\hline
\textit{de-wiki} 
& \begin{tabular}[c]{@{}l@{}}
Deutschland\textsubscript{LOC}, 
Berlin\textsubscript{LOC}, 
USA\textsubscript{LOC}, 
München\textsubscript{LOC}, \\
Frankreich\textsubscript{LOC}, 
Wien\textsubscript{LOC}, 
Zweiten Weltkrieg\textsubscript{MISC}, \\
Europa\textsubscript{LOC}, 
Österreich\textsubscript{LOC}, 
Paris\textsubscript{LOC}
\end{tabular} \\
\hline
\textit{de-tweet}
& \begin{tabular}[c]{@{}l@{}}
@SBahnBerlin\textsubscript{ORG}, 
\#S3\textsubscript{LOC}, 
\#S1\textsubscript{LOC},
\#S2\textsubscript{LOC},  
S1\textsubscript{LOC},   \\
S3\textsubscript{LOC}, 
\#S7\textsubscript{LOC}, 
Unfall\textsubscript{MISC}, 
\#S4\textsubscript{LOC}, 
\#S6\textsubscript{LOC}
\end{tabular} \\
\hline
\textit{de-news} 
& \begin{tabular}[c]{@{}l@{}}
Mark\textsubscript{MISC}, 
Frankfurt\textsubscript{LOC}, 
SPD\textsubscript{ORG}, 
CDU\textsubscript{ORG}, 
Grünen\textsubscript{ORG}, \\ 
Deutschland\textsubscript{LOC},
Sachsenhausen\textsubscript{LOC}, 
FR\textsubscript{ORG}, \\ 
Hessen\textsubscript{LOC}, 
Weltbank\textsubscript{ORG}
\end{tabular} \\
\end{tabular}
}
\caption{Top 10 frequent entities (cased) in the five normalized Bavarian and German (sub-)datasets.}
\label{tab:frequent-entities}
\end{table}

\subsection{Qualitative Observations}\label{subsec:comparison-qualitative-observations}

This section provides qualitative observations from our annotators when comparing Bavarian NER annotation to German. 
Person entities in Bavarian are the most distinct from German.
Firstly, family names in Bavarian usually come before given names \citep[p.~71]{weiss1998syntax}, e.g., \textit{Dreßen} is the family name in \textit{Dreßen Thomas}. 
Secondly, as diminutives are used more frequently in Bavarian, many given names have specific shortened versions as nicknames \citep[p.~108]{merkle1993bairische}, e.g. \textit{Sebastian} becomes \textit{Basti} or \textit{Wastl.}
Furthermore, given names are typically preceded by definite articles \citep[p.~69--71]{weiss1998syntax}, e.g.,
\textit{d'Maria} and \textit{da Michel}.

We also observe distinctions in the genitive case marking between the two dialects. 
Bavarian has three common grammatical cases: nominative, accusative, and dative. 
The genitive determiner in German is replaced by the combination of the preposition \textit{vo} `from' and a dative determiner in Bavarian (\citealt[p.~96]{merkle1993bairische},
\citealt{bulow_structures_2021}).
For example, `Association of National Olympic Committees' is translated as  \textit{Vaeinigung vo de Nationoin Olympischn Komitees} in Bavarian with \textit{vo de}, rather than \textit{Vereinigung der Nationalen Olympischen Komitees} in German with genitive \textit{der}.

To summarize, \S\ref{sec:compare-german} illustrates substantial variations in lexical distributions, entities, and grammatical constructions captured in annotations.
These (sub-)datasets were organized to capture entities in designated topics, genres, and dialects.

\section{NER Experiments \& Results}\label{sec:ner-experiments-results}
This section examines whether neural architectures, language models, and external training resources help Bavarian NER tagging.
\S\ref{subsec:experiments-setups} explains the basic experiment setups.
\S\ref{subsec:experiments-in-domain} shows that in-domain NER tagging performances are lower in Bavarian than in German.
\S\ref{subsec:experiments-cross-domain} further illustrates degradations when testing on out-of-domain (OOD) (sub-)datasets.
\S\ref{subsec:experiments-sequential-training} and \S\ref{subsec:experiments-joint-training} show improvements through sequential and joint training with multiple Bavarian and German (sub-)datasets.
However, NER on Bavarian or tweets is still more difficult than on German or the more prescriptive wiki and news genres.

\subsection{Setups}\label{subsec:experiments-setups}

We employ MaChAmp \citep{van-der-goot-etal-2021-massive} for our NER experiments using the masked CRF decoder with BIO encoding and its default hyperparameters.
MaChAmp allows joint and sequential training and achieves satisfying performances on the MultiCoNER shared task \citep{malmasi-etal-2022-semeval, plank-2022-sliced, goot-2022-machamp}.
We use two top-performing LMs, German \textit{deepset/gbert-large}\footnote{https://huggingface.co/deepset/gbert-large} (GBERT, \citealt{chan-etal-2020-germans}) 
and multilingual \textit{xlm-roberta-large}\footnote{https://huggingface.co/xlm-roberta-large} (XLM-R, \citealt{conneau-etal-2020-unsupervised}).
Experiments are conducted on five normalized German and Bavarian (sub-)datasets introduced in \S\ref{sec:compare-german}.
Models are trained on an NVIDIA A100 GPU, and we report 3-run averages (avg) with standard deviations (std) on the Span F1 metric.

\subsection{In-domain Training}\label{subsec:experiments-in-domain}
We compare in-domain training of our Bavarian NER sub-corpora with three German ones, two from the same genres and one benchmark news corpus. 
Table~\ref{tab:experiment-in-domain-results} presents in-domain results with XLM-R and GBERT and the number of training tokens and entities in each dataset for comparison. 
NER is more difficult on \textit{bar-wiki}, \textit{bar-tweet}, and \textit{de-tweet} since they are smaller in training size and represent non-mainstream variations in dialects or genres. Overall, German \textit{news/wiki} text reaches F1 scores in the high 80s, while this drops to the 70s on the social media genre and for Bavarian.
Moreover, \textit{bar-tweet} has sparser entity density (2.23\%) than the other (sub-)datasets (see \S\ref{subsec:comparison-type-mapping-distributions}).

\begin{table}[h!bt]
\centering
\resizebox{0.5\textwidth}{!}{
\begin{tabular}{c|rrrrr}
In-domain 
& \multicolumn{1}{c}{\textit{bar-wiki}} 
& \multicolumn{1}{c}{\textit{bar-tweet}}
 & \multicolumn{1}{c}{\textit{de-wiki}} 
 & \multicolumn{1}{c}{\textit{de-tweet}} 
 &  \multicolumn{1}{c}{\textit{de-news}} \\
\hline
\#TrainToks & 61.4K  & 71.8K & 232.4K  & 47.0K  & 207.0K \\
\#TrainEnts  & 2.7K   & 1.6K & 12.9K  & 7.3K  & 10.0K \\
\%Ents/Toks & 4.40\% & 2.23\% & 5.55\% & 15.5\% & 4.83\% \\
\hline
\textbf{XLM-R}  
&  \textbf{72.91\textsubscript{0.67}}
& \textbf{77.55\textsubscript{0.64}}
& 85.67\textsubscript{0.80}  
& 77.14\textsubscript{0.69} 
& 88.35\textsubscript{0.33}  
\\
GBERT  
& 72.17\textsubscript{1.75} 
& 73.30\textsubscript{6.98} 
& \textbf{86.68\textsubscript{0.50}}
& \textbf{79.75\textsubscript{0.62}}
& \textbf{90.23\textsubscript{0.37}}
\\
\end{tabular}
}
\caption{In-domain training results (3-run avg\textsubscript{std} Span F1) and the number of training tokens and entities in five Bavarian and German (sub-)datasets.}
\label{tab:experiment-in-domain-results}
\end{table}

\noindent
In \S\ref{subsec:experiments-cross-domain}-\S\ref{subsec:experiments-joint-training}, we experiment with cross-domain testing and sequential and joint training to optimize NER tagging on Bavarian.
We choose XLM-R for further experiments given its higher Span F1s and lower standard deviations on Bavarian.

\subsection{Cross-domain Evaluation}\label{subsec:experiments-cross-domain}

We conduct out-of-domain testing across five \makebox{(sub-)}datasets to assess whether and to what extent distinctions among training data surface in model performance.
Table~\ref{tab:experiment-out-of-domain-results} presents results trained on the (sub-)datasets in the left columns and tested on the top rows. 
We include \underline{in-domain testing} in the diagonal for comparison, and the best out-of-domain scenario on the target dataset is \textbf{bolded}.

\begin{table}[h!bt]
\centering
\resizebox{0.5\textwidth}{!}{
\begin{tabular}{c|rrrrr}
Train\textbackslash Test 
& \multicolumn{1}{c}{\textit{bar-wiki}} 
& \multicolumn{1}{c}{\textit{bar-tweet}}
 & \multicolumn{1}{c}{\textit{de-wiki}} 
 & \multicolumn{1}{c}{\textit{de-tweet}} 
 &  \multicolumn{1}{c}{\textit{de-news}} \\
\hline
\textit{bar-wiki} 
& \underline{72.91\textsubscript{0.67}}
& 35.33\textsubscript{2.37}
& 71.62\textsubscript{1.35}
& 55.39\textsubscript{2.07}
& 71.39\textsubscript{1.39}
\\
\textit{bar-tweet} 
& \textbf{60.07\textsubscript{3.50}}
& \underline{77.55\textsubscript{0.64}}
& 66.39\textsubscript{1.92}
& 48.63\textsubscript{2.70}
& 67.43\textsubscript{2.89}
\\
\textit{de-wiki} 
& 57.45\textsubscript{5.29}
& 24.07\textsubscript{1.45}
& \underline{85.67\textsubscript{0.80}}
& \textbf{60.21\textsubscript{3.82}}
& \textbf{81.91\textsubscript{2.42}}
\\
\textit{de-tweet}
& 51.32\textsubscript{7.34}
& \textbf{37.32\textsubscript{2.23}}
& 65.40\textsubscript{1.25}
& \underline{77.14\textsubscript{0.69}}
& 67.59\textsubscript{1.56}
\\
\textit{de-news}
& 48.73\textsubscript{4.98}
& 29.31\textsubscript{2.49}
& \textbf{78.66\textsubscript{0.24}}
& 58.54\textsubscript{1.83}
& \underline{88.35\textsubscript{0.33}}
\\
\end{tabular}
}
\caption{Out-of-domain (OOD) evaluation results (3-run avg\textsubscript{std} Span F1) across five (sub-)datasets.}
\label{tab:experiment-out-of-domain-results}
\end{table}

Firstly, models trained on the larger \textit{de-wiki} and \textit{de-news} (sub-)corpora perform badly on Bavarian, 15.46 and 24.18 percentage points lower than in-domain performances on \textit{bar-wiki}, 53.48 and 48.24 lower on \textit{bar-tweet}.
This shows that a model trained on the smaller but targeted Bavarian data (\textit{de-wiki} and \textit{de-news} are three times bigger) is more beneficial.
\textit{de-wiki} and \textit{de-news} models also lose 10+ percentage points on \textit{de-tweet}, but cross-genre degradations between \textit{de-wiki} and \textit{de-news} are relatively small.
On the other hand, models trained on Bavarian also perform poorly on German data.
These low cross-domain performances substantiate dissimilarities across Bavarian and German and the need for our NER annotations on Bavarian.

\subsection{Sequential Training}\label{subsec:experiments-sequential-training}

Since \textit{bar-wiki} and \textit{bar-tweet} are smaller and do not target specific topics, we suffer from fewer training tokens and entities (see Table~\ref{tab:experiment-in-domain-results}). 
We experiment with sequential training to provide access to larger external entity resources.
Table~\ref{tab:experiment-sequential-results} demonstrates results where we first train on the (sub-)datasets in the left column and then continue training and evaluating on the top row target ones.

\begin{table}[h!bt]
\centering
\resizebox{0.5\textwidth}{!}{
\begin{tabular}{c|lllll}
First\textbackslash Target
& \multicolumn{1}{c}{\textit{bar-wiki}} 
& \multicolumn{1}{c}{\textit{bar-tweet}}
 & \multicolumn{1}{c}{\textit{de-wiki}} 
 & \multicolumn{1}{c}{\textit{de-tweet}} 
 &  \multicolumn{1}{c}{\textit{de-news}} \\
\hline
\textit{bar-wiki} 
& \multicolumn{1}{c}{/}
& \textbf{79.27\textsubscript{1.33}$\uparrow$}
& 85.64\textsubscript{0.52}
& \textbf{77.26\textsubscript{0.83}$\uparrow$}
& 88.50\textsubscript{0.54}$\uparrow$
\\
\textit{bar-tweet} 
& 68.86\textsubscript{3.28}
& \multicolumn{1}{c}{/}
& 85.57\textsubscript{0.16}
& 76.48\textsubscript{1.42}
& 88.84\textsubscript{0.45}$\uparrow$
\\
\textit{de-wiki} 
& \uwave{\textbf{73.67\textsubscript{1.16}$\uparrow$}}
& 78.07\textsubscript{1.00}$\uparrow$
& \multicolumn{1}{c}{/}
& 76.76\textsubscript{0.81}
& 88.57\textsubscript{0.44}$\uparrow$
\\
\textit{de-tweet}
& 69.55\textsubscript{2.53}
& 76.32\textsubscript{0.82}
& \textbf{86.08\textsubscript{1.14}$\uparrow$}
& \multicolumn{1}{c}{/}
& \textbf{88.89\textsubscript{0.09}$\uparrow$}
\\
\textit{de-news}
& 71.79\textsubscript{1.12}
& 78.65\textsubscript{1.87}$\uparrow$
& 85.11\textsubscript{1.28}
& 76.71\textsubscript{1.08}
& \multicolumn{1}{c}{/}
\\
\hline
In-domain
&  72.91\textsubscript{0.67}
& 77.55\textsubscript{0.64}
& 85.67\textsubscript{0.80}  
& 77.14\textsubscript{0.69} 
& 88.35\textsubscript{0.33}  
\end{tabular}
}
\caption{Sequential training results (3-run avg\textsubscript{std} Span F1) on five (sub-)datasets. $\uparrow$ indicates an increase from in-domain baseline and the highest is bolded; e.g., \uwave{\textbf{73.67\textsubscript{1.16}$\uparrow$}} is the avg\textsubscript{std} first trained on \textit{de-wiki} and then on \textit{bar-wiki} which outperforms ($\uparrow$) the in-domain baseline 72.91\textsubscript{0.67} and achieves the best performance on \textit{bar-wiki} (bolded).}
\label{tab:experiment-sequential-results}
\end{table}

\noindent
Compared to in-domain training (the bottom row in Table~\ref{tab:experiment-sequential-results}), 
first training on one other dataset and then on the target improves performance on all five (sub-)datasets. 
Early training on same-genre \textit{de-wiki} helps \textit{bar-wiki} tagging by 0.76.
\textit{bar-tweet} achieves the largest gain of 1.67 by first training on \textit{bar-wiki}, potentially since they belong to the same dialect and \textit{bar-wiki} has higher entity density. 
First training on German wiki and news also helps \textit{bar-tweet} but not on the same genre \textit{de-tweet}, narrowly focusing on transportation texts. 
On German, \textit{de-tweet}'s topic-heavy entities result in the highest performances on the much larger \textit{de-wiki} (86.08) and \textit{de-news} (88.89).
All other Bavarian and German (sub-)datasets also slightly improve the performance on the seminal but 20-year-old \textit{de-news}. 
To summarize, we show that sequential training can improve target data performance.
We hypothesize that language diversity in genres, dialects, and topic-specific vocabularies contribute to these improvements.

\subsection{Joint Training}\label{subsec:experiments-joint-training}

In addition to sequential training, 
we analyze how training jointly with all five (sub-)datasets while developing and testing on the target influences model performances. 
We also pipeline joint training with another round of training on the target dataset (joint+seq).
Table~\ref{tab:experiment-joint-results} demonstrates the results.

\begin{table}[h!bt]
\centering
\resizebox{0.5\textwidth}{!}{
\begin{tabular}{c|lllll}
 & \multicolumn{1}{c}{\textit{bar-wiki}} 
& \multicolumn{1}{c}{\textit{bar-tweet}}
 & \multicolumn{1}{c}{\textit{de-wiki}} 
 & \multicolumn{1}{c}{\textit{de-tweet}} 
 &  \multicolumn{1}{c}{\textit{de-news}} \\
\hline
joint 
& 81.73\textsubscript{1.52}$\uparrow$	
& \textbf{78.17\textsubscript{0.91}$\uparrow$}
& 	\textbf{85.89\textsubscript{0.22}$\uparrow$}
& 	76.58\textsubscript{0.41}	
& 	87.62\textsubscript{0.46}
\\
joint+seq 
& \textbf{84.09\textsubscript{2.92}$\uparrow$}
& 77.80\textsubscript{0.62}$\uparrow$
& 	85.17\textsubscript{0.21}
& 	75.78\textsubscript{1.08}
& 	\textbf{88.67\textsubscript{0.85}$\uparrow$}
\\
\hline
In-domain
&  72.91\textsubscript{0.67}
& 77.55\textsubscript{0.64}
& 85.67\textsubscript{0.80}
& 77.14\textsubscript{0.69} 
& 88.35\textsubscript{0.33}  
\end{tabular}
}
\caption{Joint training results (3-run avg\textsubscript{std} Span F1) on five (sub-)datasets with or without sequential training on the target dataset.}
\label{tab:experiment-joint-results}
\end{table}

\noindent
Joint training on five (sub-)datasets considerably improves performance on \textit{bar-wiki} (+8.82) and moderately on \textit{bar-tweet} (+0.62), but not as much for the three German datasets.
When adding pipeline training on the target, 
\textit{bar-tweet} increases by another 2.36.
Adding sequential to joint training also improves the large \textit{de-news} benchmark.

To summarize, we here (\S\ref{sec:ner-experiments-results}) present four experiment scenarios, in-domain, cross-domain, sequential, and joint training, on five (sub-)datasets.
We first establish in-domain NER baselines on \corpus{}.
Secondly, we demonstrate the unsatisfying performances of directly applying German-trained models to Bavarian data, substantiating the necessity of our sizeable Bavarian annotations.
Thirdly, we demonstrate first training on other (sub-)datasets and then training on the target one improves model performance. 
This could result from a larger training size, higher entity density, or more topic diversity. 
Lastly, jointly train on all five (sub-)datasets and then on the target achieves radical enhancement +11.18 on \textit{bar-wiki} NER tagging. 

\section{Multi-Task Learning with Dialect Identification}\label{sec:multi-task-dialect-identification}

Dialect identification (DID) is an NLP task discriminating between similar dialects.
German is well-known for its intrinsic dialectal
variation, but existing German DID datasets are geographically labeled and centralized on Swiss German, e.g., the VarDial 2017-2019 shared tasks \citep{zampieri-etal-2017-findings, zampieri-etal-2018-language, zampieri-etal-2019-report}.
Given our gold selection of Bavarian wiki and tweet texts, this section conducts multi-task learning (MTL) between NER tagging and DID for Bavarian and German. 


We conduct separate DID experiments for tweets and wikis.
Our tweet DID dataset consists of all \textit{bar-tweet} sentences (75K tokens) and an equal amount of German tweets from an in-house archive.\footnote{We refrain from using MobIE for tweet DID due to its topic specificity.} 
For wiki DID, we extract 75K tokens from Bavarian and German wiki pages.
We run the MaChAmp classification task with XLM-R and report averages on Micro F1 in the top-left corner of Table~\ref{tab:experiment-multi-task-dialect-identification}.

\begin{table*}[t!bh]
\centering
\resizebox{0.90\textwidth}{!}{
\begin{tabular}{c|cc|ccccc}
& \multicolumn{2}{c|}{DID} & \multicolumn{5}{c}{NER} \\
 & \multicolumn{1}{c}{\textit{wiki}} 
 & \multicolumn{1}{c|}{\textit{tweet}}
 & \multicolumn{1}{c}{\textit{bar-wiki}} 
& \multicolumn{1}{c}{\textit{bar-tweet}}
 & \multicolumn{1}{c}{\textit{de-wiki}} 
 & \multicolumn{1}{c}{\textit{de-tweet}} 
 &  \multicolumn{1}{c}{\textit{de-news}} \\

\hline
in-domain
& \textbf{99.93\textsubscript{0.06}} 
& 96.17\textsubscript{1.10} 
&  72.91\textsubscript{0.67}
& 77.55\textsubscript{0.64}
& \textbf{85.67\textsubscript{0.80}}
& \textbf{77.14\textsubscript{0.69}}
& \textbf{88.35\textsubscript{0.33}}  
\\
multi-task
& 99.85\textsubscript{0.06}	
& \textbf{96.60\textsubscript{0.07}} 
& \textbf{84.17\textsubscript{2.07}}
& \textbf{78.28\textsubscript{1.54}}
& 	85.45\textsubscript{0.34}
& 	74.88\textsubscript{1.11}
& 	88.01\textsubscript{0.42} \\
\hline
\begin{tabular}[c]{@{}c@{}}Best \\ Model \\ Improvement \end{tabular}
&  \begin{tabular}[c]{@{}c@{}}in-domain \\  99.93\textsubscript{0.06} \\ / \end{tabular}
&  \begin{tabular}[c]{@{}c@{}}multi-task \\ 96.60\textsubscript{0.07} \\ 
+0.43
\end{tabular}
&  \begin{tabular}[c]{@{}c@{}}multi-task \\  84.17\textsubscript{2.07} \\ 
+11.26
\end{tabular}
&  \begin{tabular}[c]{@{}c@{}} seq-bar-wiki \\ 79.27\textsubscript{1.33}  \\ 
+1.72 
\end{tabular}
&  \begin{tabular}[c]{@{}c@{}} seq-de-tweet \\ 86.08\textsubscript{1.14} \\ 
+0.41  
\end{tabular}
&  \begin{tabular}[c]{@{}c@{}} seq-bar-wiki \\ 77.26\textsubscript{0.83}  \\ 
+0.12
\end{tabular}
&  \begin{tabular}[c]{@{}c@{}} seq-de-tweet \\ 88.89\textsubscript{0.09} \\ 
+0.54
\end{tabular}
\end{tabular}
}
\caption{Performances on binary dialect identification (DID) between Bavarian and German, with multi-task learning (MTL) combining 5 NER and 2 DID (sub-)datasets, and our best setups (3-run avg\textsubscript{std}).}
\label{tab:experiment-multi-task-dialect-identification}
\end{table*}

Models yield commendable performances on both DID datasets.
Still, the average F1 score on tweet DID is 3.76 lower than on wiki. 
This results from the frequent code-switching between Bavarian and German in tweets and their shorter sentences, i.e., 10.64 tokens/sentence (incl. @mentions and \#hashtags) in tweet DID data but 14.94 in wiki. 

Furthermore, we evaluate multi-task learning with equal weights on the 5 NER and 2 DID tasks. 
As shown in Table~\ref{tab:experiment-multi-task-dialect-identification}, MTL is beneficial for our non-canonical data: It enhances both DID on the harder tweet genre and NER on two Bavarian sub-corpora but not on wiki DID or German NER.

Lastly, we summarize the best-performing models throughout \S\ref{sec:ner-experiments-results}-\S\ref{sec:multi-task-dialect-identification} on the seven tasks at the bottom of Table~\ref{tab:experiment-multi-task-dialect-identification}.
Results show that MTL scores state-of-the-art on tweet DID and \textit{bar-wiki} NER; the latter improvement is particularly rewarding. 
Still, most tasks, i.e., NER on the other four (sub-)datasets, achieve the best performances via first training on an external dataset, \textit{bar-wiki} or \textit{de-tweet}, and then on the target.
We evince the advantages of data diversity in dialects, genres, and topics on dialect identification and named entity recognition.

\section{Error Analysis}\label{sec:error-analysis}
We perform error analyses on \corpus{} tagging to interpret how sequential, joint, and multi-task learning models improve over the in-domain baseline.
Table~\ref{tab:error_analysis} presents NER tagging errors from four aspects: orthography and word choices of Bavarian proper and common nouns, guidelines regarding emojis and hashtags, and faithfulness to the CoNLL 2006 rules. 

\begin{table}[h!bt]
\centering
\resizebox{0.5\textwidth}{!}{
\begin{tabular}{cc|cccc}
Example & Trans/Expl & base & seq & joint & mtl \\

\hline
\multicolumn{6}{c}{Bavarian: proper nouns} \\
 \textit{[Traunviatl]\textsubscript{LOC}} & (district name)
&  \ourcross &  \ourcross & \ourcheck & \ourcross \\
\textit{[Nordkar]\textsubscript{LOC}} & (a ski resort)
  & \ourcross   & \ourcross  & \ourcheck  & \ourcheck \\
 \textit{[Wastl]\textsubscript{PER}} & Sebastian
 & \ourcross & \ourcross &  \ourcheck  &  \ourcheck  \\
\begin{tabular}[c]{@{}c@{}}\textit{[Bechtolsheim} \\ \textit{Andy]\textsubscript{PER}}\end{tabular} & \begin{tabular}[c]{@{}c@{}}Andy \\ Bechtolsheim\end{tabular}
&  \ourcheck  &  \ourcheck  &  \ourcheck & \ourcross \\

\hline
 \multicolumn{6}{c}{Bavarian: common nouns} \\
\textit{Haisl} & house
 & \ourcross & \ourcross &  \ourcheck  &  \ourcheck  \\
\textit{Bazi} & rascal
&  \ourcross &  \ourcross & \ourcross & \ourcheck \\

\hline
 \multicolumn{6}{c}{Guideline: emoji \& hashtags in tweets} \\
\textit{[\emojigermany]\textsubscript{LOC}} & (Germany flag)
&  \ourcross &  \ourcheck & \ourcheck & \ourcheck \\
\textit{[\#minga]\textsubscript{LOC}} & \#munich
&  \ourcross &  \ourcross & \ourcheck & \ourcross \\

\hline
\multicolumn{6}{c}{Guideline: faithful to CoNLL 2006} \\
\textit{Eiropa-Zentrale} & Europe-center  
&  \ourcross &  \ourcross & \ourcheck & \ourcheck \\
\begin{tabular}[c]{@{}c@{}}\textit{[Prommenade]\textsubscript{LOC}} \\ \textit{23}\end{tabular} 
& (street name)
&  \ourcross &  \ourcross & \ourcheck & \ourcheck \\
\begin{tabular}[c]{@{}c@{}}\textit{`Dr. [Karl Ritter} \\ \textit{von Görner]\textsubscript{PER}}'\end{tabular} & (person name)
& \ourcross & \ourcross & \ourcheck  & \ourcheck    \\
\begin{tabular}[c]{@{}c@{}}\textit{``[Eazheazogtum} \\ \textit{Owaöstareich ob} \\ \textit{da Enns]\textsubscript{ORG}''}\end{tabular} &
\begin{tabular}[c]{@{}c@{}}Archduchy of \\ Austria above \\ the Enns\end{tabular}
 &  \ourcross &  \ourcheck & \ourcheck & \ourcheck \\
 

\end{tabular}
}
\caption{NER error analysis: examples of gold entity annotations and whether our in-domain baseline (base), sequential (seq), joint, and multi-task (mtl) models tag them correctly (yes \ourcheck or no \ourcross).}
\label{tab:error_analysis}
\end{table}

\corpus{} includes many new proper names such as regional locations in Bavarian or Austria (\textit{Nordkar} and \textit{Traunviatl}), person names (\textit{Verdi}), and the tradition of spelling the family name first (\textit{Bechtolsheim} is the family name in \textit{Bechtolsheim Andy}) opposite from German or English.
Unseen spellings of common nouns (\textit{Haisl} `house' or \textit{Bazi} `rascal') also result in model errors wrongly predicting them as entities.
Unlike English, all German nouns, including common nouns, are capitalized.
This makes distinguishing common and proper nouns more difficult than in English. 
Additionally, tweet texts are more colloquial, and common and proper nouns are frequently written in lowercase.
As a result, the base and sequential models perform similarly badly in these two scenarios.
Models improve on Bavarian with additional knowledge from joint and multi-task training. 

Incorporating more similarly annotated (sub-)corpora also helps adhere to the CoNLL 2006 guideline. 
In tweets, we found the joint and multi-task models more successful in tagging emojis (\emojigermany) and hashtags (\textit{\#minga}).
Across both genres, they are more faithful in excluding nominal compounds containing partly NEs (\textit{Eiropa-Zentrale}), house numbers after street names (\textit{Prommenade 23}), titles before person names (\textit{Dr. Karl Ritter Görner}), but including post-nominal prepositional phrases that specify the entity (\textit{Eazheazogtum Owaöstareich ob da Enns} `en. Archduchy of Austria above the Enns').
Owing to the multilingual language model and vast German training data, Bavarian NER can achieve high 70s to low 80s performance, not much lower than on the German datasets.

However, lexical ambiguity is an unsolved issue for NER. 
The best-performing model is still confused with polysemes such as \textit{Google}.
For example, \textit{Google} can refer to an office location (LOC), a company (ORG), or a web search product (MISC).
Similarly, whether \textit{\#zoom} refers to the online conferencing platform (MISC) or the company (ORG) is hard to distinguish, particularly with limited context.
The large portion of double-annotated documents in \corpus{} can empower future studies addressing the interplay between annotators' and models'  disagreements.

Lastly, we perform error analyses on wiki and tweet dialect identification. 
All falsely classified cases in wiki are short, i.e., titles or phrases, which provide minimal context and could be ambiguous even for human annotators.
In addition to short sentences, 
tweet DID data exhibits more code-switching and challenges model performances.
On one side, the sampled in-house German tweets contain few examples with individual Bavarian terms. 
Conversely, some \textit{bar}-labeled sentences minimally contain Bavarian. 
For example, only `\textit{Foisch}' (`wrong', \textit{Falsch} in German) is dialectal in `\textit{Lektion 3 Foisch: Da kann ich nichts machen}' (`Lesson 3 wrong: There's nothing I can do about it').
This reveals the limitations of binary sentence-level DID.
Moreover, our tweet DID annotation guideline includes predominately Bavarian sentences as \textit{bar} to upscale data coverage for NER but decreases DID performance on tweets. 
More granular token-level annotations on code-mixing and NEs \citep{solorio-etal-2014-overview} could be a promising future direction.


\section{Conclusion}\label{sec:conclusion}
This paper presents \corpus{}, a medium-sized, manually annotated named entity corpus for Bavarian Wikipedia and tweets.
We show quantitatively and qualitatively lexical and entity-level distinctions between German and Bavarian and how they affect NER performance. 
Our observations reveal the necessity to call for more dialectal datasets.
In addition to in-domain experiments, we establish state-of-the-art results on \corpus{} by incorporating German NER corpora through sequential and joint training and multi-task learning with Bavarian-German dialect identification.
Moreover, we also observe German performance gains by integrating diversity in domains -- genres, topics, and dialects. 

Future research directions include
aligning between mainstream languages and local dialects, 
studying fine-grained sub-dialectal (sub-regional) variations, 
and comparing translation-based and transfer-based approaches to dialectal NLP.
Syntactic and discourse annotations on Bavarian and other dialects are also along the line \citep{blaschke2024maibaam}.
Our dialectal NER and DID annotations can further benefit spoken corpora, for example, identifying NEs and classifying dialect regions in transcriptions of spoken data, e.g., whether and which part(s) of an interview involves dialectal speakers. 
All in all, we sincerely hope our study inspires more work on providing low-resource dialectal corpora in the future \citep{blaschke-etal-2023-survey} and quantifying their distinctiveness from higher-resourced standard language corpora.

\section*{Limitation}\label{sec:limitation}
Firstly, we acknowledge that accessing previously existing and future tweet data from Twitter (X) is getting increasingly challenging and unfavorable. However, X is a trendy social media platform with a large sample of colloquial and user-generated Bavarian texts complementing well-edited Wikipedia data. We will ensure we safely store raw and annotated data and distribute them properly to researchers. 

Secondly, we did not collect full metadata informing the sub-variety of these Bavarian texts. 
Unfortunately, location-related metadata for tweets is no longer accessible to us, and such information is frequently missing from Wikipedia articles. 
We also did not collect demographic metadata such as age and gender of users.
However, by inspecting a small sample, we hypothesize that the most represented dialect is Central Bavarian for both \textit{bar-wiki} and \textit{bar-tweet}.

Thirdly, manual dialect classification of German tweets is challenging. Tweets can be short in length and frequently code-mixed between English, standard German, Bavarian, and other dialects. Moreover, sub-varieties of German dialects can be vastly different, and one annotator may not be capable of identifying the tweet's exact dialect label. 
For example, a Central Bavarian speaker can judge a tweet as non-Central-Bavarian and non-standard-German. However, they might be unable to tell if the tweet is in, e.g., a North Bavarian dialect or a neighboring East Franconian one. 
For these reasons, we decided to use a simplistic approach to classify each tweet sentence as either mostly German, mostly Bavarian, mostly other languages or dialects, or unable to tell -- since dialect classification was mainly implemented to filter out purely standard German and other dialect tweets.
Nevertheless, some \corpus{} tweets are dialect-classified by multiple annotators, and such dialect label variations could benefit future work.

\section*{Acknowledgements}
We thank Mike Zhang, Rob van der Goot and Elisa Bassignana for giving feedback on earlier drafts of this paper.  
This work is supported by ERC Consolidator Grant DIALECT 101043235.

\section{Bibliographical References}
\bibliographystyle{lrec_natbib}
\bibliography{dialect-ner}


\newpage

\appendix

\section{NE Word Clouds}\label{sec:appx-word-cloud}

\begin{figure}[h!bt]
\centering
\begin{subfigure}{0.30\textwidth}
\includegraphics[width=\textwidth]{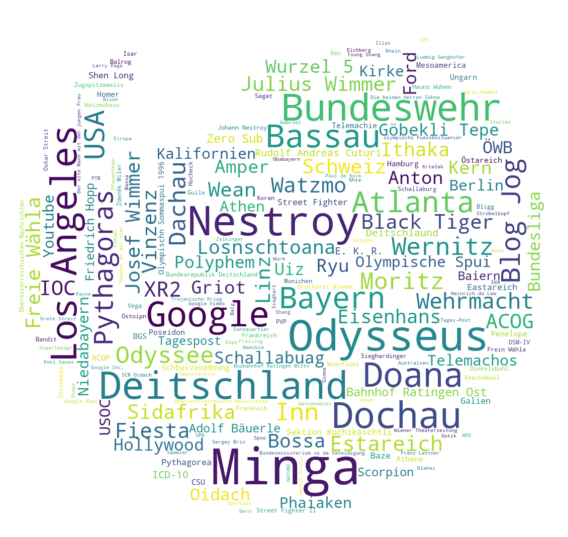}
\vspace*{-4ex}
\caption{\textit{bar-wiki}}
\label{subfig:wordclouds-bar-wiki}
\end{subfigure}
\begin{subfigure}{0.30\textwidth}
\includegraphics[width=\textwidth]{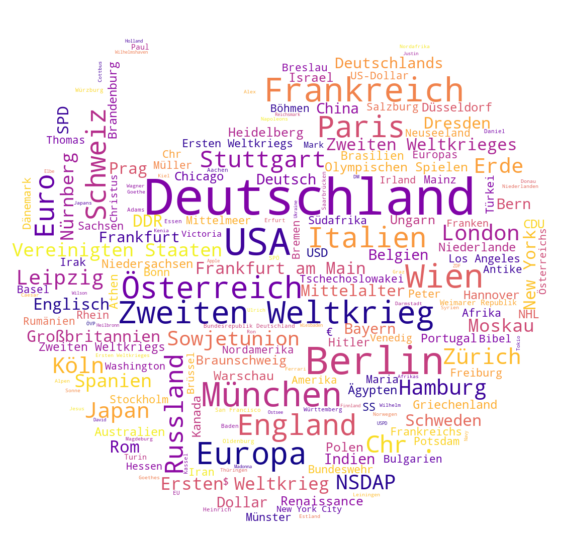}
\vspace*{-4ex}
\caption{\textit{de-wiki}}
\label{subfig:wordclouds-de-wiki}
\end{subfigure}
\begin{subfigure}{0.28\textwidth}
\includegraphics[width=\textwidth]{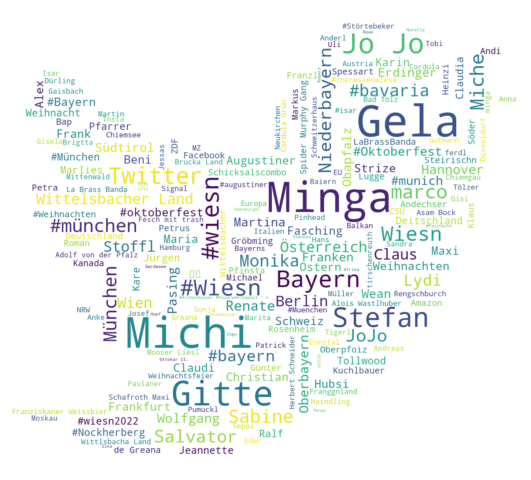}
\vspace*{-4ex}
\caption{\textit{bar-tweet}}
\label{subfig:wordclouds-bar-tweet}
\end{subfigure}
\begin{subfigure}{0.30\textwidth}
\includegraphics[width=\textwidth]{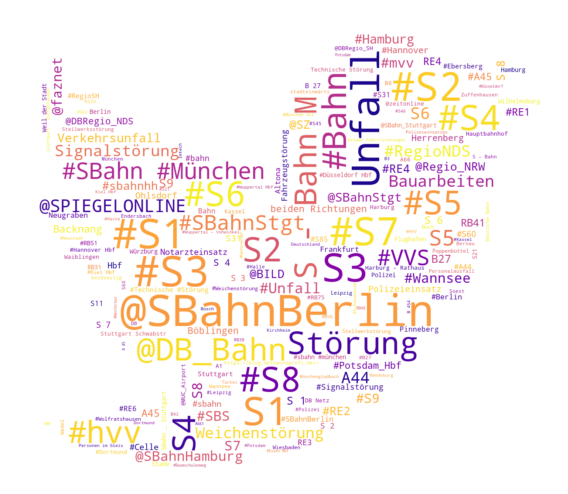}
\vspace*{-4ex}
\caption{\textit{de-tweet}}
\label{subfig:wordclouds-de-tweet}
\end{subfigure}
\begin{subfigure}{0.20\textwidth}
\includegraphics[width=\textwidth]{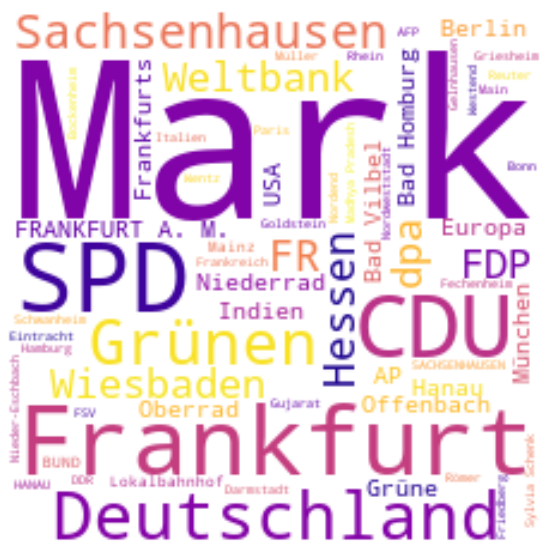}
\vspace*{-3ex}
\caption{\textit{de-news}}
\label{subfig:wordclouds-de-news}
\end{subfigure}
\caption{NE word clouds of the five datasets.}
\label{fig:wordclouds}
\end{figure}

\section{Data Statement for \corpus{}}\label{sec:appx-data-statement}

We follow \citet{bender-friedman-2018-data} to include a data statement to alleviate potential ethical issues raised during \corpus{} data collection. 

\paragraph{Header}
\begin{itemize}
\item 
\textsl{Dataset Title:}
\corpus{}

\item 
\textsl{Dataset Curator(s):}
Coordinators of this project, Siyao Peng, Verena Blaschke, Ekaterina Artemova and Barbara Plank, are academic employees at LMU Munich.
Our three annotators, Zihang Sun, Huangyan Shan, and Marie Kolm, are master students at LMU Munich. 

\item 
\textsl{Dataset Version:}
V1.0 as of \today

\item 
\textsl{Dataset Citation:}
\corpus{} should be cited by citing this publication.

\item 
\textsl{Data Statement Authors:}
Same as the authors of this publication.

\item 
\textsl{Data Statement Version:}
V1.0 as of \today

\item 
\textsl{Data Statement Citation and DOI:}
To cite this data statement, please cite this publication.

\item 
\textsl{Links to versions of this data statement in other languages:}
None
\end{itemize}

\paragraph{Executive Summary}
\corpus{} is the first named entity (NE) corpus for Bavarian German manually annotated by three annotators. 
The corpus includes 161K+ tokens and 6.6K+ entities balanced across two genres: Wikipedia articles and Twitter (X).
We annotate both coarse-grained 
person (PER), location (LOC), organization (ORG), and miscellaneous (MISC) NEs following CoNLL 2006 German guidelines \citep{tjong-kim-sang-de-meulder-2003-introduction}, 
and fine-grained derived and partially contained NEs adapted from GermEval 2014 \citep{reimers2014germeval}. 
Half of \corpus{} is double-annotated with 83+ inter-annotator agreement on typed F1. 

\paragraph{Curation Rationale}
The crucial research question behind \corpus{} and this paper is to examine whether and to what extent named entities surface differently between standard and dialectal language variations. 
We take Bavarian German as the dialectal counterpart and compare it with mainstream (high) German. 
We also contrast NEs in two representative genres: iteratively-refined Wikipedia texts versus spontaneously-iterated tweets. 
NE annotations follow the conventional BIO-encoding at the token level in both genres.
Nevertheless, Wikipedia articles are selections of continuous segments averaging over 1k+ tokens per document, whereas tweets are naturally much shorter instances in tokens.

\paragraph{Documentation for Source Datasets}
\corpus{} is directly sourced from raw Wikipedia and Twitter texts, not built on a pre-existing corpus.

\paragraph{Language Varieties}
Our dataset focuses on the Bavarian dialectal variant of German, i.e., ISO: 639-3; code: \textit{bar}; referred to as \textit{Bairisch} in standard German and \textit{Boarisch} in Bavarian German.
Even though we did not collect sub-dialectal metadata for \corpus{}, we can still claim that
the most represented sub-dialect is Central Bavarian
by inspecting a small sample.

\paragraph{Speaker Demographic}
We could not obtain full demographic information regarding Wikipedia editors and tweet users.

\paragraph{Annotator Demographic}
All three annotators are master students in their 20s, one male and two female. 
One annotator is a native Bavarian speaker, and the other two are native Mandarin Chinese speakers who majored in German during their bachelor studies and thus have full professional proficiency in German.

\paragraph{Speech Situation and Text Characteristics}
Wikipedia articles and tweets were collected between February and May 2023. 
Both are asynchronously written data directed at the general public.
However, Wikipedia is more scripted, whereas Twitter simulates more spontaneous speech.
Topic interests of the annotators contributed to the selection of Wikipedia articles.
No topic restriction was imposed while scrapping Twitter posts.

\paragraph{Preprocessing and Data Formatting}
Twitter data were scrapped using Twitter API\footnote{\url{https://developer.twitter.com/en/docs/twitter-api}} conditioned on the list of Bavarian-speaking users, and Wikipedia texts are manually copy-pasted from webpages containing continuous text segments starting from the beginning of the page. 
All texts are automatically tokenized using SoMaJo's \citep{proisl-uhrig-2016-somajo} \textit{de\_CMC} model.
Annotations are released in the CoNLL-styled tab-separated (\textit{tsv}) format where a line is 1) a token and its tab-joined BIO-encoded NE tag, 2) hashtag-initial metadata information, or 3) a blank line separating sentences. 
Most NE annotations are done using local text editors except for the last two weeks, that was piloted on a newly released Eevee annotation tool\footnote{\url{https://axelsorensen.github.io/EeveeTest/}} \citep{sorensen2024eevee}.
For Twitter data, we anonymize mentions to @Mention but allow NE annotations on \#Hashtags and emojis.

\paragraph{Capture Quality}
We ensure the dialectal authenticity of \textit{bar-tweet} by asking annotators to label each tweet sentence whether they are mostly Bavarian, German, another language or dialect, or unintelligible. 
We only keep mostly Bavarian tweets in our data release.

\paragraph{Limitations}
See the Limitation section of this paper.

\paragraph{Metadata}

\begin{itemize}
\item 
\textsl{License:} 
CC-BY 4.0.\footnote{\url{https://creativecommons.org/licenses/by/4.0/deed.en}}

\item
\textsl{Annotation Guidelines:} 
\url{https://github.com/mainlp/BarNER/blob/main/MaiNLP_NER_Annotation_Guidelines.pdf}

\item 
\textsl{Annotation Process:} 
The annotators were hired and compensated for their
work following national salary rates.

\item 
\textsl{Dataset Quality Metrics:} 
Cohen's kappa.

\item 
\textsl{Errata:}
None so far. Please report errors by contacting the authors or opening an issue at \url{https://github.com/mainlp/BarNER/}.
\end{itemize}

\paragraph{Disclosures and Ethical Review}
This work
is supported by ERC Consolidator Grant DIALECT
101043235.

\paragraph{Other}
None.

\paragraph{Glossary}
\begin{itemize}
  \setlength{\itemsep}{0pt}
\item PER: Person
\item LOC: Location
\item ORG: Organization
\item MISC: Miscellaneous
\item LANG: Language
\item RELIGION: Religion
\item EVENT: Event
\item WOA: work-of-art
\item -part: partly containing a nominal NE
\item -deriv: morphologically derived from a nominal NE
\end{itemize}


\end{document}